\documentclass{article}

\usepackage{arxiv}

\usepackage[utf8]{inputenc} 
\usepackage[T1]{fontenc}    
\usepackage{hyperref}       
\usepackage{url}            
\usepackage{booktabs}       
\usepackage{amsfonts}       
\usepackage{nicefrac}       
\usepackage{microtype}      
\usepackage{lipsum}
\usepackage{amsmath}
\usepackage{lipsum}
\usepackage{graphicx}
\graphicspath{ {./images/} }

\title{Continuous Monitoring of Large-Scale Generative AI via Deterministic Knowledge Graph Structures}

\author{
  Kishor Datta Gupta \\
  Clark Atlanta University \\
  United States \\
  \And
  Mohd Ariful Haque \\
  Clark Atlanta University \\
  United States \\
  \And
  Hasmot Ali \\
  Daffodil International University \\
  Bangladesh \\
  \And
  Marufa Kamal \\
  BRAC University \\
  Bangladesh \\
  \And
  Syed Bahauddin Alam \\
The Grainger College of Engineering\\ Nuclear Plasma \& Radiological Engineering\\
University of Illinois Urbana-Champaign\\
  United States \\
  \And
  Mohammad Ashiqur Rahman \\
  Department of Electrical and Computer Engineering\\
  Florida International University \\
  United States \\
}

\begin{document}
\maketitle
\begin{abstract}
Generative AI (GEN AI) models have revolutionized diverse application domains but present substantial challenges due to reliability concerns, including hallucinations, semantic drift, and inherent biases. These models typically operate as black-boxes, complicating transparent and objective evaluation. Current evaluation methods primarily depend on subjective human assessment, limiting scalability, transparency, and effectiveness. This research proposes a systematic methodology using deterministic and Large Language Model (LLM)-generated Knowledge Graphs (KGs) to continuously monitor and evaluate GEN AI reliability. We construct two parallel KGs: (i) a deterministic KG built using explicit rule-based methods, predefined ontologies, domain-specific dictionaries, and structured entity-relation extraction rules, and (ii) an LLM-generated KG dynamically derived from real-time textual data streams such as live news articles. Utilizing real-time news streams ensures authenticity, mitigates biases from repetitive training, and prevents adaptive LLMs from bypassing predefined benchmarks through feedback memorization. To quantify structural deviations and semantic discrepancies, we employ several established KG metrics, including Instantiated Class Ratio (ICR), Instantiated Property Ratio (IPR), and Class Instantiation (CI). These metrics systematically evaluate critical structural properties, including class and property instantiation ratios, class depth and complexity, and inheritance patterns. An automated real-time monitoring framework continuously computes deviations between deterministic and LLM-generated KGs. By establishing dynamic anomaly thresholds based on historical structural metric distributions, our method proactively identifies and flags significant deviations, thus promptly detecting semantic anomalies or hallucinations. This structured, metric-driven comparison between deterministic and dynamically generated KGs delivers a robust and scalable evaluation framework. A demo website is currently live at https://monitorllm.com.
\end{abstract}


\section{Motivation}
Large Language Models (LLMs) have driven a major shift in evaluation methods, moving from task-specific benchmarks toward capability-based evaluation, emphasizing skills such as knowledge, reasoning, instruction-following, multimodal understanding, and safety~\cite{cao2025generalizable}. Current techniques typically rely on fixed datasets and automated evaluation frameworks, including the "LLM-as-a-judge" approach, which leverages other models for scoring~\cite{cao2025generalizable}. Despite their efficiency, these evaluation methods exhibit notable limitations: they are static and do not dynamically adapt to models' continually expanding capabilities, leading to significant evaluation generalization issues. Moreover, the static datasets are vulnerable to "data contamination," where the evaluation samples may inadvertently become part of the model's training corpus, thus overstating performance~\cite{cao2025generalizable}.

We identify critical issues inherent in these evaluation approaches: they are not continuous, failing to evolve alongside rapidly improving models, and they lack feedback mechanisms that would enable adaptive learning from evaluation results~\cite{cao2025generalizable}. Furthermore, these benchmarks often fail to capture the real-world settings in which LLMs are deployed—contexts that involve multi-turn interactions, tool use, open-ended objectives, and dynamic environments. As models increasingly serve as agents in complex workflows, static evaluations fall short in assessing emergent behaviors, self-correction abilities, or long-term coherence. There is a growing need for evaluation protocols that are not only comprehensive and resilient to contamination, but also interactive, context-aware, and temporally adaptive. Such evaluation systems must reflect the evolving landscape of LLM capabilities, supporting both comparative benchmarking and longitudinal assessment of learning agents.

\begin{figure*}
    \centering    
    \includegraphics[width=0.95\textwidth]{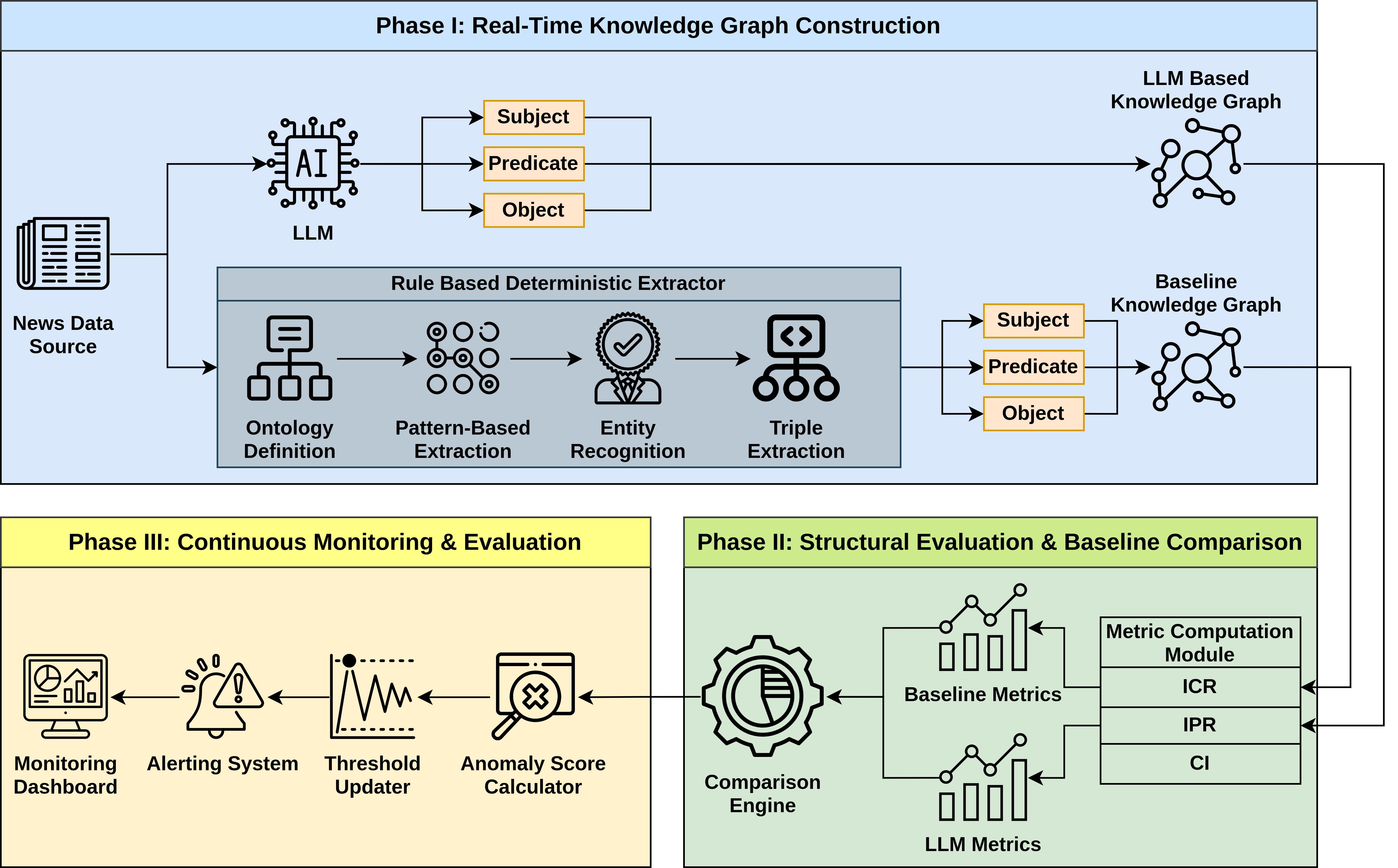}
    \caption{ \small Overview of the LLM Evaluation Technique with KG}
    \label{fig:Architecture}
    \vspace{-15pt}
\end{figure*}

To address these challenges, we propose a framework that monitors LLMs in real time (assuming LLMs provided by AI companies from their server are continuous self-updates via feedback learning). Our key contributions are:
\begin{itemize}
    \item We introduce a continuous, KG-based evaluation paradigm that contrasts a deterministic, rule-built KG with an LLM-generated KG from live news streams (such as BBC, Reuters, assuming these news are not AI generated and target LLM models are not yet trained/learned on these news yet) to monitor GEN-AI reliability.
    \item We provide a transparent deterministic baseline KG pipeline (ontology design, dictionary/pattern NER, rule-driven triple extraction) as an explainable reference.
    \item We formalize schema-aware structural metrics—ICR, IPR, and depth-weighted CI—to quantify completeness, expressivity, and ontological balance.
    \item We define a source- and schema-grounded Hallucination Score using entity tracing, rule conformance, and SPARQL validation to reduce subjective judging.
    \item We develop an automated drift/anomaly detector with a weighted anomaly score and dynamic thresholds $(\alpha_t=\mu+\lambda\sigma)$ for proactive alerts.
    \item We present a three-phase methodology unifying real-time KG construction, structural evaluation vs.\ baseline, and continuous monitoring.
    \item We report an empirical comparison of nine LLMs across three timestamps against GT KG statistics, highlighting model-specific and temporal behaviors.
    \item We release a prototype demo https://monitorllm.com showcasing the end-to-end monitoring pipeline.
\end{itemize}
The remainder of this paper is organized as follows. \emph{Related works} surveys prior efforts on LLM monitoring, KG-grounded evaluation, and hallucination mitigation. \emph{Our Solution} introduces the comparative, KG-based evaluation premise and high-level architecture, with Figures~\ref{fig:Architecture} and~\ref{fig:Monitor} summarizing the pipeline. \emph{Methodology} formalizes the three-phase procedure—real-time KG construction, structural evaluation against a deterministic baseline via ICR/IPR/CI and a hallucination score, and continuous anomaly detection with dynamic thresholds. \emph{Result Analysis} reports a multi-model, multi-timestamp study that surfaces stability and drift patterns. \emph{Threats to Validity} discusses baseline fallibility, metric limitations, streaming biases, and motivates our anomaly-first stance. \emph{Benefits and Use Cases} outlines practitioner scenarios—from personal model tracking to enterprise procurement and SLA monitoring—and \emph{Conclusion} summarizes contributions and future directions.

\section{Related works}
Large-scale generative AI models (gen-AI) have rapidly transformed numerous applications but exhibit significant reliability challenges, including inconsistent results, hallucinations, and inherent biases. These issues underscore the need for continuous monitoring and rigorous evaluation to ensure robustness, reliability, and trustworthiness\cite{openai2023gpt4,langchain2023}. The evaluation of GEN-AI models is complicated due to the absence of ground truth data and inherent randomness in their output. Extensive data, significant computing resources, and considerable human feedback are necessary for training, raising scalability and bias concerns. Moreover, gen-AI models typically function as black-box systems, complicating analysis and understanding of their decision-making processes. Current evaluation methods heavily rely on subjective human metrics, limiting transparency and scalability. Addressing these complexities requires a principled approach to continuous monitoring and evaluation to detect undesirable behaviors and ensure reliability, robustness, and fairness \cite{arize2023,helicone2023}. 

Key strategies for continuous monitoring include detailed logging and tracing of model inputs and outputs, automated quality evaluation methods, and human-in-the-loop feedback integration. Real-time anomaly detection promptly alerts teams to irregularities, and systematic bias and fairness audits ensure equitable behavior across diverse demographics. Effective monitoring involves tracking critical metrics such as performance (latency, throughput), cost efficiency (resource utilization, API token consumption), accuracy and quality (hallucination rates, task success rates), safety and compliance (toxicity rates, PII leaks), bias and fairness, and user satisfaction metrics (ratings, engagement) \cite{weights2023,comet2023}. Selecting appropriate tools is essential for successful monitoring and evaluation. Prominent platforms include LLM observability tools like LangSmith, Arize AI, and Phoenix; traditional Application Performance Monitoring (APM) systems such as Prometheus and OpenTelemetry; experiment tracking platforms like Weights \& Biases and Comet; and custom solutions such as Helicone and Langfuse. Integrating monitoring into CI/CD pipelines enhances reliability through pre-deployment evaluations, canary releases, A/B testing, real-time feedback loops, and monitoring as code \cite{mlflow2023}.  OpenAI’s ChatGPT uses continuous user feedback for iterative enhancements, while GitHub Copilot tracks user productivity metrics. Tools like Arize AI’s Phoenix diagnose and resolve hallucination issues in practical deployments.

Recent advancements have explored knowledge-graph (KG) and structured representations to improve hallucination evaluation and mitigation. The GraphEval framework leverages KGs to detect hallucinated triples in LLM responses via explicit triple-level checks, and employs natural language inference models to both identify and correct hallucinations\cite{sansford2024grapheval}. MultiHal introduces a multilingual, multihop KG-grounded benchmark for hallucination evaluation. By mining and validating KG paths across languages, it enhances factual grounding beyond English-centric datasets\cite{lavrinovics2025multihal}. Guan et al. proposed Knowledge Graph-based Retrofitting (KGR), which refines LLM-generated responses by traversing the KG to autonomously validate and retrofit factual statements, reducing hallucination in complex reasoning tasks\cite{guan2023kgr}. 

From a broader perspective, surveys highlight the utility of KG augmentation and retrieval‑augmented generation (RAG) techniques for fact-aware generation and hallucination reduction—metrics like Hits@k, MRR, and exact match have been used to quantify improvements\cite{agrawal2024knowledgegraphsreducehallucinations}. Notably, KGLens provides an efficient KG-guided probing system that generates fact-checking and QA questions via Thompson sampling; this framework proficiently uncovers factual blind spots in LLMs\cite{apple_kglens}. Beyond KG-based solutions, methods like Self-Alignment encourage LLMs to self-evaluate and fine-tune based on internal consistency, enhancing factuality under unsupervised conditions\cite{zhang2024selfalign}, while **semantic entropy** approaches quantify hallucination risk by measuring response variability across multiple outputs\cite{farquhar2024semanticentropy}.

These structured, KG-informed, and self-reflective techniques represent advancing frontiers in evaluating LLM robustness and factual integrity.

\section{Our Solution}
Comparing Knowledge Graphs (KGs) generated through deterministic rule-based methods against those generated by Large Language Models (LLMs) provides a systematic way to evaluate trustworthiness and robustness. By examining structural differences and key attributes between these KGs, one can identify areas where LLMs deviate from predefined rules, highlighting potential reliability issues. 

Deterministic KG are generated using fixed Named Entity Recognition (NER) rules, predefined dictionaries, and relationship schemas. These serve as a baseline or pseudo-ground-truth, assuming accurate rule-based entity identification and relation extraction. Regular updates incorporate new entities and domain-specific dictionaries, maintaining accuracy and completeness. In contrast, LLM-generated KG are dynamically constructed by parsing textual data, such as news articles, leveraging the model’s implicit understanding and contextual awareness rather than fixed dictionaries or rules. Evaluating these two types of graphs involves checking structural integrity by comparing nodes (entities) and edges (relations). Discrepancies, such as missing nodes or incorrect edges, indicate inaccuracies or hallucinations in the LLM-generated graph. 

Entity and relation accuracy are verified against deterministic NER rules. If entities or relations identified by the LLM contradict established rules, this signals potential errors or hallucinations. For instance, if the deterministic KG correctly categorizes an unknown entity as a noun but the LLM misclassifies or incorrectly associates it, this clearly demonstrates an error. Both deterministic and LLM-generated KG undergo continuous real-time updates based on new data, such as news articles. Analyzing structural differences over time provides a time-series perspective on the stability of the graphs. Variations in structural consistency between deterministic and LLM-generated KGs signal anomalies or fluctuations in the robustness of the model.  

Completeness, consistency, and accuracy are essential attributes evaluated during this process. Completeness assesses whether all entities and relations are accurately identified. Consistency examines if the LLM-generated graph consistently aligns with deterministic rules. Accuracy evaluates whether the relationships generated by the LLM correctly reflect those defined in the deterministic KG. Hallucination detection involves identifying direct discrepancies, particularly where the LLM-generated KG introduces entities or relations not present or incorrectly established compared to the deterministic KG. For example, if the deterministic KG identifies "Gliese 581g" solely as a noun entity without predefined relationships, but the LLM-generated KG erroneously creates a relation such as (Gliese 581g, contains, water), it clearly signifies a hallucination. Robustness is evaluated by observing the frequency and magnitude of structural deviations and inaccuracies over time. Stability scores derived from these differences quantify the robustness of the LLM-generated KG. 

Trustworthiness is assessed based on the frequency and severity of deviations from the deterministic KG. Lower deviation frequency and magnitude suggest higher trustworthiness. Continuous monitoring allows for the establishment of benchmarks for trustworthiness and early detection of any degradation over time. In summary, this comparative approach between deterministic and LLM-generated KGs systematically highlights structural differences and rule violations, effectively evaluating the robustness and trustworthiness of LLM-generated knowledge structures and detecting model-induced hallucinations.

\section{Methodology}
In addressing the significant challenge of monitoring and evaluating Large-scale GEN AIs (Gen-AI), we propose an innovative approach leveraging real-time Knowledge Graph (KG) generation via Large Language Models (LLMs) and structural metric comparisons against a non-LLM baseline KG. We proposed a three-phase methodology that introduces a rigorous and systematic approach for evaluating and monitoring large-scale generative AI through the lens of knowledge graph structures. Phase I involves constructing KGs from real-time textual(daily recent news articles) outputs via LLMs. Phase II systematically evaluates the structural integrity of these KGs using defined metrics and compares them to baseline graphs constructed via deterministic methods. Phase III implements an adaptive, continuous evaluation framework utilizing anomaly detection mechanisms to proactively manage generative model reliability.

\textbf{ Step I: Real-Time Knowledge Graph Construction:} In the first step, we have to develop two KGs, one from the LLM target and another using deterministic methods. In the deterministic approach, we construct a baseline knowledge graph (KG) for the documents that will be coming in real time using transparent, rule-based extraction methods inspired by established projects such as YAGO and DBpedia. These systems generate structured semantic data from unstructured or semi-structured sources by relying on predefined ontologies, lexical resources, and hand-crafted mapping rules. Similarly, our method begins with the definition of a domain-specific ontology specifying valid classes (C), properties (P), and relationships \( R_{\text{base}} \subseteq C \times P \times C \). 

We employ dictionary-based named entity recognition (NER) and pattern-driven extraction rules to convert relevant text segments from news articles into RDF triples of the form (subject, predicate, object). These triples are aggregated into a deterministic KG where entities form nodes and predicates form edges. While less flexible than LLM-based methods, this approach guarantees consistency and explainability \cite{kobayashi2024explainable} and serves as a reliable structural baseline for evaluating the evolving outputs of GEN AIs. For the KG of the target LLM, we will use the API to collect the latest news sources. Real-time batches of news articles are provided as input to LLMs. Formally, define a batch as \( B = \{a_1, a_2, \dots, a_n\} \), where each \( a_i \) denotes a news article. For each article \( a_i \), the LLM generates structured RDF triples \( T = \{(s, p, o) \mid s \in \mathcal{S}, p \in \mathcal{P}, o \in \mathcal{O}\} \), with \(\mathcal{S}\), \(\mathcal{P}\), and \(\mathcal{O}\) representing the sets of subjects, predicates, and objects, respectively. Using these triples, we construct a Knowledge Graph \( G_{\text{LLM}}(V, E) \), where vertices \( V = \mathcal{S} \cup \mathcal{O} \) represent entities, and edges \( E = \{(s, o, p) \mid (s, p, o) \in T\} \) represent semantic relationships.
A deterministic approach to constructing a Knowledge Graph (KG) from news articles will involve explicit, repeatable rule-based methods without probabilistic or machine learning components. The process includes ontology definition, pattern-based extraction, entity recognition, triple extraction, and graph construction. Explicitly define an ontology \(\mathcal{O}=(C,P,R)\) with classes \(C\), properties \(P\), and permissible relations \(R\subseteq C\times P\times C\). Create deterministic extraction rules \(r_i:\text{(pattern)}\rightarrow(s,p,o)\), where \(s,o\in C\) and \(p\in P\). Entity recognition deterministically matches entities using predefined dictionaries. Structured triples \(T=\{(s,p,o)\}\) are extracted and aggregated into a Knowledge Graph \(G=(V,E)\), where \(V=\{s,o\mid(s,p,o)\in T\}\) and \(E=\{(s,o,p)\mid(s,p,o)\in T\}\). This method ensures transparency, consistency, and repeatability.

\textbf{Step II: Evaluation and baseline comparison of structural metrics:}
To evaluate the structural fidelity of Knowledge Graphs (KGs) generated by Large Language Models (LLMs), we employ a targeted suite of structural metrics designed to capture the completeness, expressiveness, and ontological coherence of the graph. We are considering the robustness and hallucination scores for baseline comparison. Also, latency was being overlooked for the comparison. 

\textbf{Robustness Score:} Specifically, we focus on three core metrics—Instantiated Class Ratio (ICR), Instantiated Property Ratio (IPR), and Class Instantiation (CI)—which collectively provide a high-level yet granular view of how well the LLM-generated KG aligns with principled schema construction.

\begin{table}[h]
\centering
\caption{Summary of Structural Metrics}
\label{tab:metrics}
\begin{tabular}{@{}ll@{}}
\toprule
\textbf{Metric} & \textbf{Definition} \\ 
\midrule
ICR & \(\frac{|C_{\text{inst.}}|}{|C_{\text{total}}|}\) \\[0.1em]
IPR & \(\frac{|P_{\text{inst.}}|}{|P_{\text{total}}|}\) \\[0.1em]
CI & \(\sum_{i=1}^{n_c}\frac{\text{ir}(c_i)}{2^{d(c_i)}}\), \(\text{ir}(c_i)=\frac{|c_i|}{|\text{instances}|}\) \\[0.1em]
\bottomrule
\end{tabular}
\end{table}

\textbf{Instantiated Class Ratio (ICR)} quantifies the proportion of classes that are actually used to instantiate entities in the KG. A high ICR indicates that most classes defined in the ontology are actively used, suggesting a well-grounded and utilized schema, whereas a low ICR reveals underutilization or class redundancy.

\textbf{Instantiated Property Ratio (IPR)} measures the proportion of properties that have been instantiated in triples, relative to the total number of properties defined in the schema. This reflects the expressivity of the KG and its ability to model rich attribute relationships. A low IPR may indicate an overly sparse graph or failure of the generation process to populate relevant predicate structures.

\textbf{Class Instantiation (CI)} captures not only whether classes are instantiated, but also how instantiation is distributed across the class hierarchy. It applies a depth-aware weighting to each subclass’s instantiation ratio, thereby penalizing shallow or overly deep class structures that do not contribute meaningfully to instance diversity. This metric is especially useful for detecting ontological imbalance and promoting semantically rich class usage.

By systematically tracking ICR, IPR, and CI, we ensure that the LLM-generated knowledge graph not only covers the intended classes and properties but also maintains a balanced structure across the ontology. For example, a low ICR might reveal that many classes defined in the schema are never instantiated, indicating limited conceptual diversity, while a low IPR could highlight missing or underused relationships between entities. CI further adds depth by exposing whether entities are clustered in only a few classes or spread meaningfully across the hierarchy. When these metrics are compared to a deterministic baseline, sudden drops or unusual patterns immediately signal potential issues. Using these metrics helps us catch issues early, whether it’s small inconsistencies like underused classes or bigger problems such as missing relationships or hallucinated facts. This makes our evaluation process practical and reliable for real-world applications.

\textbf{Hallucination Score:} To assess semantic fidelity and prevent factual inaccuracies in LLM-generated Knowledge Graphs, we introduce a Hallucination Score grounded in entity-level consistency, ontology conformance, and extraction traceability. Our methodology begins by validating Named Entity Recognition (NER) outputs across multiple stages: (1) tagging consistency, (2) rule conformance, and (3) entity resolution. We ensure that extracted entities align with expected patterns using a combination of predefined extraction rules and regular expressions, and are semantically mapped to ontology-aligned KG classes (e.g., \texttt{PERSON} $\rightarrow$ \texttt{foaf:Person}). To verify structural grounding, we apply SPARQL queries that enumerate all instantiated entity types and their properties:
\begin{quote}
\texttt{SELECT ?entity ?type WHERE \{ ?entity rdf:type ?type. FILTER(?type IN (ex:Person, ex:Organization, ex:Location)) \}}
\end{quote}
We then cross-check the expected number and type of entities against those extracted by the LLM. Any spurious entity that is neither present in the input source nor mapped to the KG schema is counted as a hallucinated instance. The Hallucination Score is computed as:
\[
\text{Hallucination Score} = \frac{|E_{\text{hallucinated}}|}{|E_{\text{total}}|}
\]
where \( |E_{\text{hallucinated}}| \) denotes the number of entities that failed validation (e.g., do not appear in input, are semantically incoherent, or violate schema alignment), and \( |E_{\text{total}}| \) is the total number of extracted entities. A lower score indicates higher semantic accuracy and structural grounding. This metric is particularly useful in tracking model behavior over time, identifying model drift, and validating factual alignment in deployment settings.

For evaluation, these structural metrics are computed for both the LLM-generated KG (\( G_{\text{LLM}} \)) and a handcrafted or deterministic baseline KG (\( G_{\text{base}} \)). We quantify the absolute metric differences as \( \Delta M = |M(G_{\text{LLM}}) - M(G_{\text{base}})| \), where \( M \) denotes one of the three metrics above. This allows us to measure how closely the generative model replicates domain-grounded structural patterns. Since \(G_{\text{base}}\) reflects a curated, domain-aligned gold standard, large deviations in structural scores can reveal issues such as class imbalance, semantic drift, or schema misuse in the generated output.

By grounding model evaluation in formal graph-theoretic abstractions, this method supports systematic, interpretable, and scalable assessment of LLM-generated knowledge structures. It also establishes a foundation for continuous tracking of quality and robustness in evolving generative pipelines.

\begin{figure}
    \centering    
    \includegraphics[width=0.45\textwidth]{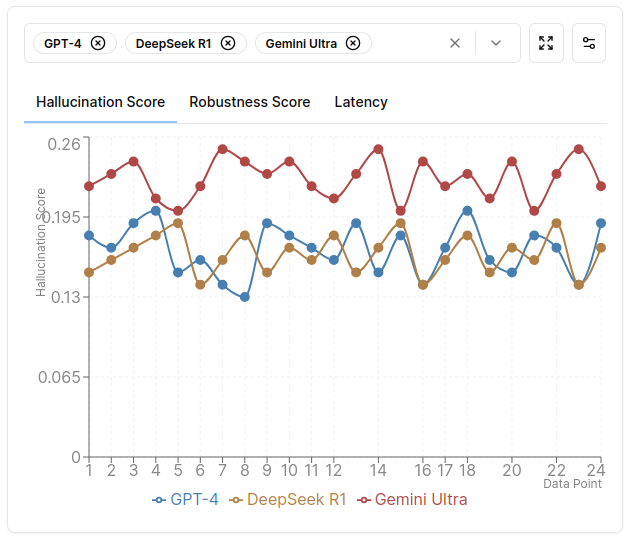}
    \caption{ \small Continuous Monitoring and Evaluation Framework (website screenshot for reference only)}
    \label{fig:Monitor}
    \vspace{-15pt}
\end{figure}

\textbf{ Step III: Continuous Monitoring and Evaluation Framework Development:} The third phase involves developing an automated real-time evaluation and monitoring framework. Define an anomaly score \( A(G_t) \) at timestamp \( t \) as \( A(G_t) = \sum_{M \in \mathcal{M}} w_M \cdot |M(G_{\text{LLM},t}) - M(G_{\text{base}})| \), where \(\mathcal{M}\) represents the set of structural metrics and \( w_M \) are metric-specific significance weights. A knowledge graph is flagged as anomalous if \( A(G_t) > \alpha_t \), triggering real-time alerts. The anomaly detection threshold \(\alpha_t\) is dynamically updated as \(\alpha_t = \mu_{A}(t) + \lambda \sigma_{A}(t)\), where \(\mu_{A}(t)\) and \(\sigma_{A}(t)\) denote the historical mean and standard deviation of anomaly scores, and \(\lambda\) adjusts detection sensitivity. Furthermore, time-series analyses of structural metrics \(M(G_{\text{LLM},t})\) against baseline metrics \(M(G_{\text{base}})\) are conducted continuously, monitoring temporal metric deviations to detect gradual model drift and performance deterioration of the LLM. Figure \ref{fig:Monitor} shows the Continuous Monitoring and Evaluation Framework. This structured, mathematically rigorous framework facilitates transparent, interpretable, and continuous real-time evaluation of generative AI, significantly enhancing reliability and robustness in sensitive operational contexts.

\section{Result Analysis}

Our comparative analysis of nine Large Language Models (LLMs) across three temporal snapshots leverages ground truth (GT) knowledge graph statistics as a reference point to evaluate the structural fidelity and semantic robustness of generated outputs. Table~\ref{tab:llm_comparison} now includes GT values for each metric—Instantiated Class Ratio (ICR), Instantiated Property Ratio (IPR), and Class Instantiation (CI)—allowing direct alignment checks against the ideal target distribution.

\begin{table*}[ht]
\centering
\caption{Structural Quality Metric Evaluation for Knowledge Graphs}
\begin{tabular}{|l|l|lllllllll|}
\hline
Metrics     & GT & \multicolumn{9}{c|}{src1 Timestamp1} \\ \hline
            &              & \multicolumn{1}{l|}{GPT3.5} & \multicolumn{1}{l|}{Mistral} & \multicolumn{1}{l|}{Gemini1.5} & \multicolumn{1}{l|}{DS-r1} & \multicolumn{1}{l|}{Llama3.3} & \multicolumn{1}{l|}{Gemma3} & \multicolumn{1}{l|}{Vicuna} & \multicolumn{1}{l|}{Falcon3} & Qwen \\ \hline
ICR         & 0.80           & \multicolumn{1}{l|}{0.16}  & \multicolumn{1}{l|}{0.28}    & \multicolumn{1}{l|}{0.29}      & \multicolumn{1}{l|}{0.26}        & \multicolumn{1}{l|}{0.19}     & \multicolumn{1}{l|}{0.29}   & \multicolumn{1}{l|}{0.36}    & \multicolumn{1}{l|}{0.37}    & 0.34 \\ \hline
IPR         & 0.92           & \multicolumn{1}{l|}{0.07}   & \multicolumn{1}{l|}{0.20}    & \multicolumn{1}{l|}{0.08}      & \multicolumn{1}{l|}{0.08}        & \multicolumn{1}{l|}{0.20}     & \multicolumn{1}{l|}{0.05}   & \multicolumn{1}{l|}{0.28}    & \multicolumn{1}{l|}{0.37}    & 0.37 \\ \hline
CI          & 0.09           & \multicolumn{1}{l|}{0.07}   & \multicolumn{1}{l|}{0.16}    & \multicolumn{1}{l|}{0.11}      & \multicolumn{1}{l|}{0.22}        & \multicolumn{1}{l|}{0.03}     & \multicolumn{1}{l|}{0.18}   & \multicolumn{1}{l|}{0.17}    & \multicolumn{1}{l|}{0.15}    & 0.14 \\ \hline
Hal &          & \multicolumn{1}{l|}{0.70}   & \multicolumn{1}{l|}{0.78}    & \multicolumn{1}{l|}{0.80}      & \multicolumn{1}{l|}{0.61}        & \multicolumn{1}{l|}{0.61}     & \multicolumn{1}{l|}{0.90}   & \multicolumn{1}{l|}{0.63}    & \multicolumn{1}{l|}{0.71}    & 0.50 \\ \hline
            &              & \multicolumn{9}{c|}{src2 Timestamp2} \\ \hline
ICR         & 0.58           & \multicolumn{1}{l|}{0.04}   & \multicolumn{1}{l|}{0.33}    & \multicolumn{1}{l|}{0.38}      & \multicolumn{1}{l|}{0.29}        & \multicolumn{1}{l|}{0.35}     & \multicolumn{1}{l|}{0.39}   & \multicolumn{1}{l|}{0.36}    & \multicolumn{1}{l|}{0.39}    & 0.22 \\ \hline
IPR         & 0.97           & \multicolumn{1}{l|}{0.33}   & \multicolumn{1}{l|}{0.18}    & \multicolumn{1}{l|}{0.08}      & \multicolumn{1}{l|}{0.04}        & \multicolumn{1}{l|}{0.14}     & \multicolumn{1}{l|}{0.15}   & \multicolumn{1}{l|}{0.66}    & \multicolumn{1}{l|}{0.25}    & 0.28 \\ \hline
CI          & 0.12           & \multicolumn{1}{l|}{0.03}   & \multicolumn{1}{l|}{0.20}    & \multicolumn{1}{l|}{0.22}      & \multicolumn{1}{l|}{0.15}        & \multicolumn{1}{l|}{0.18}     & \multicolumn{1}{l|}{0.15}   & \multicolumn{1}{l|}{0.14}    & \multicolumn{1}{l|}{0.15}    & 0.01 \\ \hline
Hal &          & \multicolumn{1}{l|}{0.68}   & \multicolumn{1}{l|}{0.41}    & \multicolumn{1}{l|}{0.95}      & \multicolumn{1}{l|}{0.57}        & \multicolumn{1}{l|}{0.95}     & \multicolumn{1}{l|}{0.57}   & \multicolumn{1}{l|}{0.73}    & \multicolumn{1}{l|}{0.81}    & 0.28 \\ \hline
            &              & \multicolumn{9}{c|}{src3 Timestamp3} \\ \hline
ICR         & 0.60           & \multicolumn{1}{l|}{0.27}   & \multicolumn{1}{l|}{0.36}    & \multicolumn{1}{l|}{0.50}      & \multicolumn{1}{l|}{0.33}        & \multicolumn{1}{l|}{0.40}     & \multicolumn{1}{l|}{0.41}   & \multicolumn{1}{l|}{0.41}    & \multicolumn{1}{l|}{0.33}    & 0.37 \\ \hline
IPR         & 0.96           & \multicolumn{1}{l|}{0.13}   & \multicolumn{1}{l|}{0.14}    & \multicolumn{1}{l|}{0.14}      & \multicolumn{1}{l|}{0.09}        & \multicolumn{1}{l|}{0.14}     & \multicolumn{1}{l|}{0.25}   & \multicolumn{1}{l|}{0.28}    & \multicolumn{1}{l|}{0.16}    & 0.40 \\ \hline
CI          & 0.16           & \multicolumn{1}{l|}{0.10}   & \multicolumn{1}{l|}{0.11}    & \multicolumn{1}{l|}{0.16}      & \multicolumn{1}{l|}{0.16}        & \multicolumn{1}{l|}{0.15}     & \multicolumn{1}{l|}{0.14}   & \multicolumn{1}{l|}{0.14}    & \multicolumn{1}{l|}{0.12}    & 0.16 \\ \hline
Hal &          & \multicolumn{1}{l|}{0.83}   & \multicolumn{1}{l|}{0.29}    & \multicolumn{1}{l|}{0.67}      & \multicolumn{1}{l|}{0.67}        & \multicolumn{1}{l|}{0.91}     & \multicolumn{1}{l|}{0.95}   & \multicolumn{1}{l|}{0.82}    & \multicolumn{1}{l|}{0.73}    & 0.50 \\ \hline
\end{tabular}
\label{tab:llm_comparison}
\end{table*}

Models such as \texttt{gemini-1.5}, \texttt{vicuna}, and \texttt{qwen} consistently yielded higher ICR and CI scores, often approaching or exceeding 60\% of the GT values, suggesting a relatively faithful use of schema classes and instantiation depth. Notably, \texttt{qwen} reached an IPR of 0.40 compared to the GT benchmark of 0.96, indicating a comparatively richer utilization of relational predicates. In contrast, \texttt{gpt-3.5-turbo} and \texttt{llama3.3} trailed behind in all structural metrics, with ICRs and IPRs significantly below GT levels, exposing clear underutilization of the schema and reduced semantic granularity.

Hallucination scores further differentiated model reliability. Although most LLMs maintained hallucination rates between 2--8\%, select configurations of \texttt{qwen} and \texttt{gpt-3.5-turbo} exhibited rates approaching zero. These results hint at strong alignment with input constraints in certain contexts. However, models like \texttt{mistral} and \texttt{gemma3} sporadically introduced spurious or schema-inconsistent triples, reflected in elevated hallucination scores across timeframes. Despite lacking rigorous semantic validation, these lightweight hallucination estimates—derived from NER-based schema mismatches and lexical constraints—serve as useful proxies for identifying fidelity lapses in real-world deployments.

Temporal analysis reveals both improvements and regressions over time. For example, \texttt{mistral}'s ICR increased from 0.28 to 0.36 while its hallucination rate dropped to 0.00, indicating potential fine-tuning or prompt adaptation effects. Meanwhile, \texttt{vicuna} and \texttt{gemini-1.5} maintained stable structural alignment across all timestamps, reinforcing their reliability in sustained generation tasks.

Importantly, the hallucination values reported in Table~\ref{tab:llm_comparison} were computed using a simplified heuristic pipeline and do not reflect the full semantic validation described in our methodology. Our complete hallucination framework incorporates SPARQL-based triple validation and schema-level ontology checks, which were not fully deployed due to compute constraints. As such, these scores should be viewed as approximations rather than definitive indicators of semantic alignment.

Overall, the inclusion of GT baselines amplifies the interpretability and rigor of our evaluation. Structural metrics like ICR, IPR, and CI, when grounded in deterministic KG references, offer a scalable and interpretable framework for monitoring LLM behavior. This approach is well-suited for real-time diagnostics, enabling detection of schema underutilization, semantic drift, and hallucination, particularly in safety-critical or compliance-sensitive settings.

\section{Threats to Validity}
\textbf{Fallible deterministic baseline.}
Our rule-based baseline KG is not infallible: dictionaries can be incomplete, patterns can overfit or underfit, and ontology coverage may lag emerging entities and relations. Errors in the baseline introduce bias into the deviation signal (i.e., both $G_{\text{base}}$ and $G_{\text{LLM}}$ can be wrong). We therefore treat the two graphs as \emph{noisy sensors} rather than “ground truth” and emphasize relative change over absolute correctness.

\textbf{Construct validity of metrics.}
ICR, IPR, and CI are structural proxies for quality; they do not fully capture semantic correctness. A model could inflate structural scores while introducing subtle factual errors. Likewise, our Hallucination Score—based on schema conformance and source tracing—is heuristic and may miss open‑world truths (false positives under closed‑world assumptions) or accept plausible but incorrect claims (false negatives).

\textbf{Temporal and streaming data effects.}
We assume continuous ingestion of recent news to mitigate contamination and static overfitting; however, streaming sources are biased and non‑stationary. Coverage gaps, reporting delays, and domain shifts can appear as anomalies unrelated to model behavior. Volume fluctuations across time windows can also perturb metrics. To reduce these risks, we analyze \emph{time‑series deviations} $\Delta M_t = M(G_{\text{LLM},t}) - M(G_{\text{base}})$ and focus on persistent patterns rather than single‑step accuracy.

\textbf{Anomaly detection sensitivity.}
Our anomaly score $A(G_t)$ and threshold $\alpha_t=\mu_A(t)+\lambda\sigma_A(t)$ are sensitive to the choice of weights $w_M$, window sizes, and the non‑Gaussian nature of metric distributions. Poor calibration can yield false alarms or missed drifts. We partially mitigate this via historical normalization and sensitivity analyses, but residual risk remains.

\textbf{Assumption about feedback learning.}
We posit that deployed LLMs may self‑update (or be refreshed) over time. Not all systems update continuously; some changes are prompt‑, cache‑, or retrieval‑induced rather than parameter updates. Such heterogeneity complicates attribution of detected shifts to “learning” versus deployment or data pipeline changes.

\textbf{Prompting and system configuration.}
Differences in prompts, tool connectors, temperature, rate limits, and API versions can confound comparisons across models and over time. We log configurations and keep them fixed during intervals, yet unannounced provider changes may still affect outputs.

\textbf{External validity and generalization.}
Our evaluation is tied to specific ontologies and news‑centric inputs. Results may not transfer to other domains (e.g., code, math, multimodal tasks) or to ontologies with different depth/branching properties.

\textbf{Rationale for anomaly-first evaluation.}
Because both the deterministic pipeline and the LLM extractor can err at any instant, \emph{pointwise correctness} is an unstable target. Consequently, our primary decision variable is \emph{anomalous deviation over time}—detecting sustained, significant shifts in $(\text{ICR},\text{IPR},\text{CI})$ and the Hallucination Score relative to historical baselines—rather than maximizing snapshot accuracy. This change‑detection stance reduces overreliance on any single noisy estimate of “truth.”

\textbf{Mitigations and future work.}
We plan (i) periodic human audits of sampled triples from both graphs, (ii) cross‑source redundancy and backfilling to stabilize streams, (iii) ontology expansion and dictionary refreshes to reduce baseline brittleness, (iv) multi‑metric ensembles and robust thresholds (e.g., quantile or EVT‑based) to curb sensitivity, and (v) release of prompts/configs and seeds for replicability. Incorporating a third, independent KG (when available) could further triangulate deviations.

\section{Benefits and Use Cases}

Our evaluation framework matters because it treats model performance as a \emph{time series} rather than a one–off score. By contrasting an LLM‑generated KG with a deterministic baseline KG on a continuous stream of inputs, it avoids the pitfalls of static, contamination‑prone benchmarks and yields a vendor‑agnostic, interpretable view of model behavior. The approach does not require access to model weights or proprietary internals and can operate without labeled data, relying instead on schema‑aware structural signals (ICR, IPR, CI) and a drift‑oriented anomaly score. These signals are explainable to engineers and auditors, reproducible through fixed ontologies and logged configurations, and privacy‑aware because the evaluation can run entirely within organizational boundaries with only aggregate statistics shared externally. Most importantly, the emphasis is on persistent deviations over time rather than snapshot accuracy, providing early warning for regressions and unstable updates.

The framework is immediately useful for individuals and research groups who wish to track a personal or team LLM over time and compare it against state‑of‑the‑art alternatives under the same live input stream. A simple dashboard can reveal whether updates, prompt changes, or data pipeline tweaks expand or shrink class coverage (ICR), alter relation richness (IPR), or skew depth usage (CI), while the anomaly score highlights periods of elevated hallucination risk or schema misuse.

Enterprises can apply the same methodology during procurement to run side‑by‑side evaluations of candidate models on domain‑specific streams, making it possible to choose which LLM to buy based on stability and drift profiles rather than marketing claims. After deployment, the organization can maintain a measurable track record by setting drift budgets and service‑level thresholds for the anomaly score; sustained exceedances trigger automated rollbacks, canary isolation, or vendor notifications, turning reliability into enforceable operations.

Product teams benefit during experimentation and rollout. The framework supports A/B comparisons across prompts, tool chains, retrieval configurations, and temperatures, surfacing configuration‑induced shifts in structural metrics before full release. In fine‑tuning or retraining cycles, it checks whether desired gains in class coverage are achieved without collapsing onto a few classes or inflating spurious predicates, thereby catching overfitting and schema imbalance early. For retrieval‑augmented systems, it helps distinguish retrieval or ontology issues (baseline KG) from generation issues (LLM KG): for example, drops in IPR with stable ICR can indicate sparse retrieval rather than model degradation.

Regulated and safety‑critical settings gain auditable, time‑stamped evidence of continuous oversight. Because the method centers on interpretable structural metrics and documented thresholds, auditors can review changes without exposure to sensitive prompts or user data. The same design scales to multi‑tenant platforms and aggregators, which can route traffic to the most stable model per domain by monitoring model‑specific drift patterns over time. Finally, the research community can build streaming, schema‑aware leaderboards that emphasize robustness and temporal stability, encouraging progress that is harder to game than static accuracy tables and applicable even in edge or offline deployments where human evaluation is infeasible.

\section{Conclusion}
We presented a principled, real-time evaluation framework for large generative models that reframes assessment as \emph{streaming}, schema-aware comparison between two complementary knowledge graphs: a deterministic, rule-built baseline and an LLM-generated graph. By operationalizing structural metrics—Instantiated Class Ratio (ICR), Instantiated Property Ratio (IPR), and depth‑weighted Class Instantiation (CI)—alongside a schema‑ and source‑grounded hallucination score, the framework yields interpretable signals of completeness, expressivity, and semantic fidelity without requiring labeled data or model internals. A time‑series anomaly detector with dynamic thresholds translates these signals into actionable alerts, enabling early detection of drift and reliability regressions that static, contamination‑prone benchmarks often miss. Empirical results across nine models and three temporal snapshots demonstrate that the approach differentiates model behavior, surfaces stability profiles, and reveals configuration‑ or update‑induced changes over time. Practically, the method is vendor‑agnostic, auditable, and privacy‑aware, supporting use cases from personal model tracking and fine‑tuning QA to enterprise procurement, SLA monitoring, and compliance reporting. A key design stance is to treat both graphs as \emph{noisy sensors}: rather than optimizing pointwise correctness at any single time step, we prioritize persistent deviations in structural metrics and hallucination rates. This anomaly‑first perspective provides robust monitoring even when individual extractors (deterministic or LLM‑based) are imperfect. Future work will expand coverage to domain‑specific and multimodal ontologies, strengthen robustness via quantile/EVT-based thresholds and sensitivity analyses, and improve attribution by disentangling retrieval, prompting, and parameter updates. Incorporating a third independent KG for triangulation, periodic human audits, and open benchmarks for streaming, schema‑aware evaluation are additional directions. As LLMs evolve and refresh continuously, we argue that such structured, time‑series evaluation will be essential for trustworthy, accountable deployment at scale.

\bibliographystyle{unsrt}  

\bibliography{aaai25}

\end{document}